\documentclass[10pt]{article}

\usepackage[margin=1in]{geometry}
\usepackage{amsmath,amssymb,mathtools}
\usepackage{array,booktabs,float,tabularx}
\usepackage{enumitem}
\usepackage{graphicx}
\usepackage{xcolor}
\usepackage{microtype}
\usepackage[hidelinks]{hyperref}
\usepackage{natbib}
\usepackage{tikz}
\usetikzlibrary{arrows.meta,calc,fit,positioning}

\setlist{topsep=0.35em}

\newcommand{\pdt}{\textsc{PDT}}

\newcommand{\M}{\mathcal{M}}
\newcommand{\R}{\mathbb{R}}
\newcommand{\sg}{\operatorname{sg}}

\newcommand{\softmax}{\operatorname{softmax}}
\newcommand{\BCE}{\operatorname{BCE}}

\title{Parallel Decoder Transformer:\\
Planner-Conditioned Latent Coordination for\\
Model-Intrinsic Parallel Generation}

\author{Logan Robbins\thanks{Independent researcher. \texttt{ljrweb@gmail.com}}}
\date{July 2026}

\begin{document}
\maketitle

\begin{abstract}
Autoregressive language models expose one causal token frontier, even when the
requested document contains sections that could be developed concurrently.
Existing parallel-generation systems arrange external branches around an
otherwise unchanged model. We instead formulate \emph{model-intrinsic parallel
generation}: a single trained architecture owns multiple causal frontiers and
produces one next-token distribution for each frontier in every synchronized
decoding round. The Parallel Decoder Transformer (\pdt) retains a frozen
shared lower knowledge trunk and replaces the upper trunk with three
independently parameterized physical decoder stacks. A prompt-time set planner
produces three unordered continuous outlines, each hard-routed to one decoder
as persistent Plan-KV memory, while a finite product-quantized notes bus
carries block-delayed latent messages among the decoders. Autoregression is
preserved within each lane; same-round lane tokens are conditionally
independent given the source, plans, private histories, and previously
committed messages. We specify source-grounded supervision for long-form
historical exposition with single-owner cited facts and token-aligned
cross-lane dependencies, a composite objective, a staged curriculum, and
preregistered causal evaluations: plan swap and removal, delayed-message
ablation, a parameter-matched self-only control, dependency-token likelihood,
and blinded human fact audits. The architecture and evaluation pipeline are
implemented; scientific training and held-out evaluation are in progress.
This paper presents the theory, design, and falsifiable protocol, not a
positive empirical result.
\end{abstract}

\section{Introduction}
\label{sec:introduction}

A decoder-only Transformer represents one sequence distribution,
\(
p(y\mid x)=\prod_t p(y_t\mid x,y_{<t})
\),
and therefore exposes one next-token decision at a time. That factorization is
appropriate when the next passage depends on all preceding prose. It is
unnecessarily restrictive when the response decomposes into complementary
sections whose local prose is sequential but whose development is partly
concurrent, as in long-form histories, operating procedures, and technical
briefings.

Prior systems exploit such structure from outside the model.
Skeleton-of-Thought elicits a textual skeleton and expands its points through
parallel calls or batched branches \citep{ning2024sot}; APAR trains a model to
emit fork control tokens and restricts attention over the resulting paragraph
tree \citep{liu2024apar}; PASTA learns an annotation language with
asynchronous promises and explicit synchronization \citep{jin2025promise}.
These systems show that generation need not follow the final reading order.
They leave open the architectural question studied here: can the model itself
contain several persistent, physically distinct causal frontiers, learn which
part each frontier should write, and exchange only a bounded latent summary of
completed work?

We call this setting \emph{model-intrinsic parallel generation}. ``Intrinsic''
does not mean computation is free or that concurrency guarantees a wall-clock
speedup. It means that planning, semantic assignment, frontier state, and
cross-frontier communication are learned components of one model rather than a
controller issuing independent language-model requests: the model returns
three next-token distributions in one grouped forward operation, and each
frontier owns private upper-layer parameters and KV state, so decoder identity
is a physical axis rather than a natural-language role label.

The design makes four contributions.

\begin{enumerate}[leftmargin=*,itemsep=0.25em]
    \item A causal factorization for three synchronized, lane-local
    autoregressive sequences: same-round tokens never observe one another, and
    cross-lane influence flows only through messages committed at an earlier
    block boundary.

    \item An architecture realizing that factorization: a frozen shared lower
    knowledge trunk, a continuous unordered set planner, and three
    independently parameterized upper decoder stacks whose private caches and
    persistent hard-routed Plan-KV make assignment structural and
    inspectable.

    \item A finite delayed communication channel: every completed 32-token
    lane block may publish four 8-bit product-VQ indices. Producer and lag
    headers address the message; the learned code tuple is the entire
    payload.

    \item Source-grounded data, losses, curriculum, controls, and causal
    gates that separate four questions: can the executor follow oracle plans,
    does the planner learn decomposition, does plan assignment cause fact
    ownership, and do sibling messages help exactly where delayed
    dependencies require them.
\end{enumerate}

\section{Problem Formulation}
\label{sec:formulation}

\subsection{Three sections, not three guesses}

Let \(x\) contain a source document and an expository request. The required
output is a tuple of three multi-paragraph sections
\(
Y=(y^{1},y^{2},y^{3})
\)
and a presentation permutation \(\pi\in S_3\); the reader sees
\(
y^{\pi_1}\Vert y^{\pi_2}\Vert y^{\pi_3}
\).
There is no fourth synthesis decoder and no serial rewriting pass: the
sections are complementary parts of one answer, not independent samples from
which a best response is selected.

The planner maps \(x\) to three semantic plans \(Z=\{z^1,z^2,z^3\}\), where
each \(z^k=(z^k_1,\ldots,z^k_N)\) is an ordered outline of at most \(N\)
nodes. The lane-level collection is an unordered set: a physical decoder has
no permanent role such as ``background'' or ``consequences,'' and
presentation order is predicted separately from physical address. Each source
fact \(f\) has exactly one owner lane \(o(f)\); another lane may be licensed
to reference it, but a fact written only by the wrong lane does not satisfy
ownership. This rules out the weak objective in which the union of three
outputs receives credit regardless of who wrote what.

\subsection{Block-delayed causal factorization}

Generation is divided into blocks of \(\tau\) lane-local tokens (canonically
\(\tau=32\)). Let \(y^k_{b,r}\) be token \(r\) of block \(b\) in lane \(k\),
let \(y^k_{<b,r}\) be all earlier tokens in that lane, and let \(\M_{<b}\) be
the messages whose publication time plus delivery delay is no later than
block \(b\); with a one-block delay, a message written after block \(b-1\)
first affects block \(b\). The model uses the factorization
\begin{equation}
\label{eq:factorization}
p_\Theta(Y,Z,\pi\mid x)
=p_\phi(Z,\pi\mid x)
\prod_{b}
\prod_{k=1}^{3}
\prod_{r=1}^{|y^k_b|}
p_{\Theta,k}\!\left(
y^k_{b,r}
\mid x,z^k,y^k_{<b,r},\M_{<b}
\right).
\end{equation}
For a fixed synchronized microstep \((b,r)\), the three current tokens are
conditionally independent given the right side of
Equation~\ref{eq:factorization}. The sections are not globally
independent---their dependence is mediated by the common source, the jointly
produced plan set, and previously committed messages---but same-round
communication is prohibited, because it would impose an order among lanes and
destroy the claimed concurrency.

\section{Parallel Decoder Transformer}
\label{sec:architecture}

\subsection{Overview and frozen shared trunk}

Figure~\ref{fig:architecture} shows the complete topology;
Figure~\ref{fig:layer-anatomy} expands one physical upper layer. A
revision-pinned decoder-only Transformer is split at layer \(L_f\). The
embedding and layers below \(L_f\), \(T_{\theta_{\mathrm{lo}}}\), form a
single frozen trunk: the common prompt is prefilled once below the fork and
the cache is expanded onto three lane rows. Layers at and above \(L_f\) are
copied into three independently trainable physical stacks, each receiving its
lane's persistent outline and the visible dynamic-note window through
separate cross-attention residuals. Parameter sharing below the fork is not a
shared hidden trajectory: once the lanes emit different tokens, their rows
pass through the same lower weights in one grouped batch but keep separate
cache rows and hidden states.

\begin{figure}[t]
\centering
\begin{tikzpicture}[
    font=\scriptsize,
    box/.style={draw, rounded corners, align=center, minimum height=9mm,
                fill=gray!7},
    lane/.style={box, minimum width=28mm, minimum height=17mm, fill=blue!7},
    memory/.style={box, text width=35mm, minimum height=14mm,
                   fill=orange!12},
    bus/.style={box, text width=35mm, minimum height=14mm, fill=green!10},
    arrow/.style={-{Latex[length=2.2mm]}, thick},
    latent/.style={-{Latex[length=2.2mm]}, thick, dashed},
    latentboth/.style={{Latex[length=2.2mm]}-{Latex[length=2.2mm]},
                       thick, dashed}
]
\node[box, minimum width=23mm] (source) at (1.15,6.0)
    {source \(x\)\\and request};
\node[box, minimum width=42mm] (lower) at (4.8,6.0)
    {frozen shared parameter trunk\\embedding and layers \(0,\ldots,L_f-1\)};
\node[box, minimum width=27mm] (fork) at (8.7,4.65)
    {one grouped call\\branch axis \([B,3,T,d_h]\)};
\node[memory] (planner) at (12.4,6.0)
    {one-time set planner\\\(3\times N\) nodes and order \(\hat\pi\)};
\node[memory] (plans) at (14.4,4.15)
    {persistent hard addresses\\
    \(\scriptstyle z^1\!\to D_1\quad
    z^2\!\to D_2\quad z^3\!\to D_3\)};

\node[lane] (d1) at (4.1,2.35)
    {\(D_1\): layers \(L_f{:}L-1\)\\
    parameter bank \(W^1\)\\private upper KV};
\node[lane] (d2) at (7.35,2.35)
    {\(D_2\): layers \(L_f{:}L-1\)\\
    parameter bank \(W^2\)\\private upper KV};
\node[lane] (d3) at (10.6,2.35)
    {\(D_3\): layers \(L_f{:}L-1\)\\
    parameter bank \(W^3\)\\private upper KV};
\node[draw, rounded corners, dashed, inner xsep=4mm, inner ysep=6mm,
      fit=(d1)(d2)(d3)] (frontier) {};

\node[bus] (bus) at (14.4,2.05)
    {addressed dynamic notes\\
    \(4\times8\)-bit indices/lane/block\\atomic commit; \(\Delta=1\)};

\node[box, minimum width=22mm] (y1) at (4.1,0.25) {section \(y^1\)};
\node[box, minimum width=22mm] (y2) at (7.35,0.25) {section \(y^2\)};
\node[box, minimum width=22mm] (y3) at (10.6,0.25) {section \(y^3\)};
\node[box, minimum width=39mm] (render) at (7.35,-1.15)
    {render in predicted order \(\hat\pi\)};

\draw[arrow] (source) -- (lower);
\draw[arrow] (lower) -- (planner);
\draw[arrow] (lower.south east) -- (fork.north west);
\draw[arrow] (fork) -- (frontier.north);
\draw[arrow] (planner.south) -- (plans.north);
\draw[latent] (plans.west) -- (frontier.north east);
\draw[latentboth] (frontier.east) -- (bus.west);

\draw[arrow] (d1) -- (y1);
\draw[arrow] (d2) -- (y2);
\draw[arrow] (d3) -- (y3);
\draw[arrow] (y1.south) |- (render.west);
\draw[arrow] (y2) -- (render);
\draw[arrow] (y3.south) |- (render.east);
\end{tikzpicture}
\caption{System topology. Solid arrows carry token states or planner
computation; dashed arrows carry persistent plan memory or block-delayed
latent messages. The three upper stacks advance in one grouped operation and
share neither base weights nor upper-layer KV caches.}
\label{fig:architecture}
\end{figure}

\subsection{Continuous set planner}

The planner runs once, before generation. Normalized prompt states
\(
H^x=T_{\theta_{\mathrm{lo}}}(x)
\)
are projected from the trunk width \(d_h\) to planner width \(d_p\), and
three groups of \(N\) learned queries cross-attend to \(H^x\) through a small
Transformer decoder:
\begin{equation}
\hat Z,\hat V,\hat s=P_\phi(H^x),\qquad
\hat Z\in\R^{3\times N\times d_p},
\end{equation}
where \(\hat V\) holds node-validity logits and \(\hat s\in\R^3\) holds
presentation scores. The three lane groups are trained as a set: the loss
enumerates all six lane permutations and differentiates only through the
minimum-cost assignment, so set-valued targets acquire no meaning from
storage order \citep{lee2019settransformer}, while node order within a lane
remains meaningful. The planner never consumes generated tokens and is never
re-invoked, so no hidden serial planner loop reintroduces
one-step-at-a-time coordination.

\subsection{Three physical decoder stacks}

The pretrained upper layers
\(
T_{\theta_{\mathrm{hi}}}^{L_f:L-1}
\)
are cloned into
\(
D_{\theta_1},D_{\theta_2},D_{\theta_3}
\).
Every self-attention projection, MLP projection, and RMS normalization weight
carries a leading physical-decoder axis,
\(W\in\R^{3\times d_{\mathrm{out}}\times d_{\mathrm{in}}}\),
whose slices are initialized identically but are distinct parameters with
independent gradients, advanced by grouped matrix multiplication. Each upper
attention layer stores private keys and values with shape
\(
[B,3,H_{\mathrm{kv}},T,d_{\mathrm{head}}]
\).
This differs materially from three prompt labels passed through a shared
decoder: the private cache axis, branch parameter slice, and hard-routed plan
memory already determine the address, so no lane needs to infer its
assignment from a tag inside a shared cache.

\subsection{Persistent hard-routed Plan-KV}

Planner nodes are projected from \(d_p\) to a 256-wide memory representation.
At every physical upper layer \(\ell\), lane \(k\) reads only its own valid
nodes:
\begin{align}
\widetilde z^k_n &= W_z z^k_n + e^{\mathrm{node}}_n,\\
A^{\ell}_{\mathrm{plan}}(h^k,z^k)
&=W^\ell_O
\softmax\!\left(
\frac{(W^\ell_Q h^k)(W^\ell_K\widetilde z^k)^\top}{\sqrt{d_a}}
\right)
W^\ell_V\widetilde z^k .
\end{align}
The result enters through a learned gated residual. Nodes remain individually
addressable as keys and values---mean pooling into a single additive plan
vector is not the canonical path---and plan memory is read-only and persists
for the complete generation.

\subsection{Finite delayed dynamic notes}
\label{sec:notes}

Static plans define intended work; dynamic notes summarize work actually
completed. At the end of block \(b\), lane \(k\)'s block-final hidden state
is projected to \(u^k_b\in\R^{256}\), split into \(M=4\) subvectors, and
quantized:
\begin{equation}
c^k_{b,m}
=\arg\min_{c\in\{1,\ldots,C\}}
\left\|u^k_{b,m}-e_{m,c}\right\|_2^2,
\qquad M=4,\quad C=256.
\end{equation}
The transmitted payload is the tuple
\(
(c^k_{b,1},\ldots,c^k_{b,4})
\),
with capacity \(M\log_2 C=32\) bits per lane block; receivers decode the
indices through the learned codebooks \citep{vandenOord2017vq}. Every note
carries non-payload headers identifying its producer and age: Shared Notes
Cross-Attention (SNC) adds learned producer, message-kind, and log-lag
embeddings before forming keys and values, and the receiver's hidden state
forms the query, so the same addressed note can be used differently by each
lane. The window retains up to 16 blocks. Writes are atomic at a block
boundary: a note published at block \(b\) is invisible during block \(b\) and
becomes readable at block \(b+1\), and all three writes commit before any
receiver begins the next block. Both Plan-KV and SNC use zero-initialized
output projections and closed residual gates, so the untrained extension
starts as the cloned pretrained decoder rather than an uncontrolled
perturbation.

\begin{figure}[t]
\centering
\begin{tikzpicture}[
    font=\scriptsize,
    base/.style={draw, rounded corners, align=center, minimum height=11mm,
                 text width=49mm, fill=blue!7},
    plan/.style={draw, rounded corners, align=center, minimum height=12mm,
                 text width=35mm, fill=orange!12},
    notes/.style={draw, rounded corners, align=center, minimum height=12mm,
                  text width=35mm, fill=green!10},
    memory/.style={draw, rounded corners, align=center, minimum height=10mm,
                   text width=25mm, fill=gray!7},
    state/.style={draw, rounded corners, align=center, minimum height=8mm,
                  minimum width=20mm, fill=gray!7},
    add/.style={circle, draw, minimum size=5.5mm, inner sep=0pt, fill=white},
    arrow/.style={-{Latex[length=2mm]}, thick},
    residual/.style={-{Latex[length=1.8mm]}, thin},
    latent/.style={-{Latex[length=2mm]}, thick, dashed},
    query/.style={-{Latex[length=1.8mm]}, thin, dotted}
]
\node[state] (hin) at (8.0,6.8)
    {\(h^{k,\ell-1}\)};
\node[base] (attn) at (8.0,5.65)
    {\textbf{branch-private Qwen attention path}\\
    RMSNorm\({}^{k,\ell}\) \(\rightarrow\) causal self-attention
    \(W^{k,\ell}_{QKVO}\) using private
    \(K^{k,\ell}_{\mathrm{tok}},V^{k,\ell}_{\mathrm{tok}}\)};
\node[add] (addattn) at (8.0,4.55) {\(\boldsymbol{+}\)};
\node[base] (mlp) at (8.0,3.45)
    {\textbf{branch-private Qwen feed-forward path}\\
    RMSNorm\({}^{k,\ell}\) \(\rightarrow\) SwiGLU MLP
    \(W^{k,\ell}_{\mathrm{MLP}}\)};
\node[add] (addmlp) at (8.0,2.35) {\(\boldsymbol{+}\)};
\node[add] (addplan) at (8.0,1.15) {\(\boldsymbol{+}\)};
\node[add] (addnotes) at (8.0,-0.25) {\(\boldsymbol{+}\)};
\node[state] (hout) at (8.0,-1.25) {\(h^{k,\ell}\)};

\node[memory] (planmem) at (1.45,1.15)
    {read-only Plan-KV\\\(z^k_{1:N}\), node mask, positions};
\node[plan] (planattn) at (5.0,1.15)
    {\textbf{shared Plan-KV interface}\\
    query from lane \(k\); keys/values only from \(z^k\)\\
    output scaled by \(\sigma(g^\ell_{\mathrm{plan}})\)};

\node[notes] (snc) at (11.2,-0.25)
    {\textbf{shared SNC interface}\\
    producer-, kind-, and lag-addressed cross-attention\\
    output scaled by \(\sigma(g^\ell_{\mathrm{notes}})\)};
\node[memory] (notemem) at (14.75,-0.25)
    {visible \(\M_{<b}\)\\at most 16 prior blocks};

\draw[arrow] (hin) -- (attn);
\draw[arrow] (attn) -- (addattn);
\draw[arrow] (addattn) -- (mlp);
\draw[arrow] (mlp) -- (addmlp);
\draw[arrow] (addmlp) -- (addplan);
\draw[arrow] (addplan) -- (addnotes);
\draw[arrow] (addnotes) -- (hout);

\draw[residual] (hin.east) -- ++(16mm,0) |- (addattn.east);
\draw[residual] (addattn.east) -- ++(28mm,0) |- (addmlp.east);

\draw[latent] (planmem) -- (planattn);
\draw[latent] (planattn) -- (addplan);
\draw[latent] (notemem) -- (snc);
\draw[latent] (snc) -- (addnotes);
\draw[query] (addmlp.west) -| (planattn.north);
\draw[query] (addplan.east) -| (snc.north);
\end{tikzpicture}
\caption{Anatomy of upper layer \(\ell\) for physical lane \(k\), repeated at
every physical upper layer. The Qwen self-attention, normalization, MLP
weights, and token KV cache are lane-private; Plan-KV and SNC have shared
interface parameters but operate on lane-private queries and structurally
addressed memories. Dotted arrows carry the lane state into the
cross-attention queries; dashed arrows carry memory inputs and gated residual
outputs.}
\label{fig:layer-anatomy}
\end{figure}

\subsection{Synchronized parallel generation}
\label{sec:inference}

Natural generation follows one canonical loop:

\begin{enumerate}[leftmargin=*,itemsep=0.2em]
    \item encode the prompt, run the planner once to resolve three valid
    plan memories and one presentation order, prefill the common prompt
    below the fork once, and expand the lower cache to three rows;

    \item at each round, execute one grouped lower-plus-upper forward with
    the assigned Plan-KV and sample one token from each active lane's
    returned distribution;

    \item when every active lane reaches a 32-token boundary, compute all
    product-VQ writes, commit them together, and only then build the next
    block's addressed note windows;

    \item a lane that emits EOS becomes logically inactive but remains a
    masked physical row in the packed cache; other lanes continue without
    falling back to separate model calls, and the finished sections are
    decoded independently and concatenated in predicted presentation order.
\end{enumerate}

Figure~\ref{fig:block-timeline} makes the synchronization and message delay
explicit: a vertical slice through the three lane rows is one grouped
decoding round, and no lane can observe a sibling's token or block-end code
from the current block.

\begin{figure}[t]
\centering
\begin{tikzpicture}[
    font=\scriptsize,
    block/.style={draw, rounded corners, minimum width=46mm,
                  minimum height=9mm, align=center, fill=blue!7},
    nextblock/.style={block, fill=blue!4},
    commit/.style={draw, rounded corners, text width=48mm,
                   minimum height=14mm, align=center, fill=green!10},
    memory/.style={draw, rounded corners, text width=54mm,
                   minimum height=14mm, align=center, fill=green!10},
    arrow/.style={-{Latex[length=2mm]}, thick},
    latent/.style={-{Latex[length=2mm]}, thick, dashed}
]
\node[font=\scriptsize\bfseries] at (3.55,5.95) {generation block \(b\)};
\node[font=\scriptsize\bfseries] at (11.15,5.95) {generation block \(b+1\)};

\node[anchor=east] at (0.85,4.85) {\(D_1\)};
\node[anchor=east] at (0.85,3.75) {\(D_2\)};
\node[anchor=east] at (0.85,2.65) {\(D_3\)};

\node[block] (b1) at (3.55,4.85)
    {\(y^1_{b,1}\;\cdots\;y^1_{b,r}\;\cdots\;y^1_{b,32}\)};
\node[block] (b2) at (3.55,3.75)
    {\(y^2_{b,1}\;\cdots\;y^2_{b,r}\;\cdots\;y^2_{b,32}\)};
\node[block] (b3) at (3.55,2.65)
    {\(y^3_{b,1}\;\cdots\;y^3_{b,r}\;\cdots\;y^3_{b,32}\)};

\node[nextblock] (n1) at (11.15,4.85)
    {\(y^1_{b+1,1}\;\cdots\;y^1_{b+1,r}\;\cdots\;y^1_{b+1,32}\)};
\node[nextblock] (n2) at (11.15,3.75)
    {\(y^2_{b+1,1}\;\cdots\;y^2_{b+1,r}\;\cdots\;y^2_{b+1,32}\)};
\node[nextblock] (n3) at (11.15,2.65)
    {\(y^3_{b+1,1}\;\cdots\;y^3_{b+1,r}\;\cdots\;y^3_{b+1,32}\)};

\draw[very thick, blue!50!black, dotted] (3.55,5.32) -- (3.55,2.18);
\node[align=center, fill=white, inner sep=1.5pt] at (3.55,6.55)
    {one grouped round \(r\): three next-token distributions};

\draw[very thick, dashed] (7.35,5.55) -- (7.35,0.05);
\node[rotate=90, fill=white, inner sep=1.5pt] at (7.35,4.25)
    {atomic block barrier};

\node[draw=none, fit=(b1)(b2)(b3)] (completed) {};
\node[draw=none, fit=(n1)(n2)(n3)] (next) {};

\node[commit] (commit) at (4.15,0.75)
    {\textbf{after all three lanes finish block \(b\)}\\
    product-VQ each block-final state\\
    atomically commit \(\{c^1_b,c^2_b,c^3_b\}\)};
\draw[latent] (completed.south) -- (commit.north);

\node[memory] (visible) at (10.95,0.75)
    {\textbf{after delivery lag \(\Delta=1\)}\\
    \(\M_{<b+1}\) contains every block-\(b\) write\\
    with producer and lag headers\\
    all upper-layer SNC modules may read it};
\draw[arrow] (commit) -- (visible);
\draw[latent] (visible.north) -- (next.south);
\end{tikzpicture}
\caption{Synchronized causal timeline for block size \(\tau=32\). Each
microstep advances all active physical lanes in one grouped call; after the
final microstep, the three product-VQ writes are committed as one atomic set,
cannot affect block \(b\), and first enter the addressed SNC windows at block
\(b+1\).}
\label{fig:block-timeline}
\end{figure}

The loop emits up to three sampled tokens per synchronization round while
preserving token-level autoregression inside every section. For final lane
lengths \(L_1,L_2,L_3\), a serial concatenated decoder requires
\(
R_{\mathrm{serial}}=\sum_k L_k
\)
decode rounds; the packed frontier has algorithmic span
\(
R_{\pdt}=\max_k L_k
\)
after planning and prefill, for a maximum span reduction of
\begin{equation}
\label{eq:span}
\rho_{\mathrm{span}}
=\frac{\sum_{k=1}^3 L_k}{\max_k L_k}
\leq 3.
\end{equation}
Equation~\ref{eq:span} is a dependency-depth statement, not a hardware speed
claim: the upper weights are triplicated, and real latency depends on
kernels, memory traffic, occupancy, cache length, and lane balance. We
therefore report synchronization depth, arithmetic work, memory footprint,
and measured wall time separately.

\section{Source-Grounded Long-Form Supervision}
\label{sec:data}

\subsection{Domain and source ingress}

Short answers and question-answer pairs do not test the mechanism: a lane can
satisfy them before persistent plans, ownership, and delayed coordination
become meaningful. The first study therefore uses historical events and
processes whose sources naturally support several long complementary
sections; every teacher section must contain 700--1,000 Qwen tokens and four
to twelve connected paragraphs, and lists, timelines, biographies as primary
topics, and catalog-like pages are excluded. Historical text also supplies
checkable temporal constraints. ``Natural'' in the data contract means that a
record's presentation order and cross-section references obey the document's
causal and expository prerequisites, not that the record proves latent
coordination.

Each source begins from one exact English Wikimedia revision, retained as an
immutable raw bundle (wikitext, rendered semantic HTML, assessments,
reference records, revision identifiers, licenses, digests). Only a derived
representation is model-visible: a strict DOM allowlist keeps the article
title, hierarchical content headings, and ordinary prose paragraphs, with
inline links contributing only display text; tables, lists, infoboxes,
figures, citation markers, navigation, and all other non-prose apparatus are
excluded, and reference identities remain audit metadata never concatenated
into the student prompt. Articles are accepted or rejected whole---never
truncated or stitched---under the preregistered gates of
Table~\ref{tab:sourcegates}.

\begin{table}[htbp]
\centering
\small
\begin{tabularx}{\textwidth}{
>{\raggedright\arraybackslash}p{0.27\textwidth}
>{\raggedright\arraybackslash}X}
\toprule
Gate & Requirement \\
\midrule
Topic and age &
English main-namespace event or process; non-redirect, non-disambiguation;
event ended at least 25 years before acquisition. \\
Assessment &
FA, GA, A, or B from a relevant history-oriented WikiProject; no citation,
neutrality, accuracy, original-research, cleanup, or hoax maintenance
template. \\
Complete cleaned body &
3,000--7,000 pinned-Qwen tokens, at least six substantive sections, at least
twelve substantive paragraphs. \\
Citations &
At least 30 inline reference occurrences, 15 distinct cited works, eight
scholarly or institutional sources; citations on at least 70\% of
substantive paragraphs; no cited work above 25\% of occurrences. \\
Split integrity &
Related pages and manually reviewed family keys assigned to one train,
validation, or test split before teacher generation. \\
\bottomrule
\end{tabularx}
\caption{Whole-article source gates for the historical corpus.}
\label{tab:sourcegates}
\end{table}

\subsection{Two-stage teacher construction}

Teacher data are produced in two schema-constrained stages.

\paragraph{Stage A: cited facts.}
For each accepted source, the teacher extracts 18--48 atomic facts spanning
at least twelve source paragraphs and four source sections. Every fact
carries an exact source quote, character offsets, paragraph ID, and local
reference IDs that must agree with the immutable source record, plus one
plausible hard negative that alters an entity, date, quantity, or relation.

\paragraph{Stage B: joint three-lane answer.}
One request receives the complete source and validated fact inventory and
returns one expository prompt, exactly three plans, all three target
sections, and their presentation order---jointly, because cross-lane
ownership and references are properties of the complete document. Every plan
has four to eight nodes specifying its expository objective, owned facts,
permitted reference facts, dependencies, and target paragraphs. Every source
fact has exactly one \textsc{owner} lane; each other lane labels it
\textsc{reference} or \textsc{absent}, an \textsc{owner} or
\textsc{reference} label requires an exact quote in the target section, and
hard negatives must remain absent everywhere. Cross-lane references are
optional: when present, a dependency names the owner plan, node, and quote
and the receiver node and quote; the owner must precede the receiver in
presentation order; and the owner quote must fall in a generation block
visible at least one block before the first receiver token. Zero-dependency
records are valid---the compiler does not invent communication to make the
notes bus look useful.

\subsection{Tokenized training record}

Retokenization uses the exact student tokenizer and appends one EOS token to
each target, so per-lane stopping is learned rather than bolted onto
inference. A frozen 1,024-dimensional BGE encoder \citep{xiao2023cpack}
embeds every true fact and hard negative and a textual form of every outline
node; a fixed Gaussian Johnson--Lindenstrauss projection maps node targets to
512 dimensions. Each record materializes the fact embeddings, padded
continuous plan targets with node-validity masks, exact fact-to-node routing,
one active outline node per 32-token block,
\textsc{owner}/\textsc{reference}/\textsc{absent} labels at the exact lane
and block where evidence occurs, exact dependency-token masks with
source-block-to-target-token edges verified against the one-block delay, and
the reader-facing presentation ranks. The first pilot targets 128 accepted
records balanced across eight historical categories, each manually inspected;
a 2,048-record corpus is the architecture gate. Four separately curated
examples currently exercise the complete schema and trainer interface as
data-contract inspection records, not a training corpus.

\section{Learning Objective}
\label{sec:objective}

\subsection{Unordered plan matching}

For example \(i\), let \(\hat z_{k,n}\) and \(\hat v_{k,n}\) be predicted
plan nodes and validity logits, and let \(z_{k,n}\) and \(v_{k,n}\) be
teacher targets. For each lane permutation \(\sigma\in S_3\), define
\begin{align}
C_i(\sigma)
={}&
\frac{
\sum_{k,n} v_{k,n}
\left[1-\cos(\hat z_{\sigma(k),n},z_{k,n})\right]
}{
\sum_{k,n} v_{k,n}
}
\nonumber\\
&+
\frac{1}{3N}\sum_{k,n}
\BCE\!\left(\hat v_{\sigma(k),n},v_{k,n}\right).
\end{align}
The minimum-cost assignment
\(
\sigma_i^\star=\arg\min_{\sigma\in S_3} C_i(\sigma)
\)
is selected discretely and detached; gradients flow through the selected
predicted nodes, and all downstream semantic axes are expressed in matched
order.

\subsection{Execution and semantic losses}

\paragraph{Lane token likelihood.}
\(\mathcal L_{\mathrm{tok}}\) is exact teacher-forced next-token cross
entropy over every valid token, including each lane's EOS; the first token of
every target block is scored from the preceding prompt- or block-final
distribution, so block packing drops no boundary likelihood.

\paragraph{Owned-fact routing.}
The route head scores every plan node against all true and hard-negative
fact embeddings; \(\mathcal L_{\mathrm{route}}\) gives equal aggregate weight
to positive owned node--fact pairs and negatives. Reference facts are not
ownership targets for this head.

\paragraph{Outline progress.}
\(\mathcal L_{\mathrm{progress}}\) is cross entropy between each block-final
hidden state and the assigned plan node whose target paragraph occupies most
of that block.

\paragraph{Fact writing.}
For every lane, target block, and fact query, a classifier predicts
\textsc{owner}, \textsc{reference}, or \textsc{absent};
\(\mathcal L_{\mathrm{write}}\) uses inverse-frequency class weights
\(
w_c=n/(3n_c)
\),
so the many absent labels cannot define a trivial optimum, and a fact written
by the wrong physical lane is penalized even if it appears somewhere in the
combined answer.

\paragraph{Dynamic-note alignment.}
\(\mathcal L_{\mathrm{note}}\) is symmetric InfoNCE at temperature 0.07
between each quantized note and its pre-quantized projection; other
lane/block writes in the packed document supply negatives.

\paragraph{Presentation order.}
Given matched lane scores \(s\) and teacher order
\(\pi=(\pi_1,\pi_2,\pi_3)\), the Plackett--Luce loss is
\begin{equation}
\mathcal L_{\mathrm{order}}
=-\log\frac{e^{s_{\pi_1}}}{\sum_j e^{s_j}}
-\log\frac{e^{s_{\pi_2}}}{e^{s_{\pi_2}}+e^{s_{\pi_3}}}.
\end{equation}

\paragraph{Finite-channel regularization.}
For continuous subvector \(u\) and selected code \(e\),
\(\mathcal L_{\mathrm{commit}}=\|u-\sg[e]\|_2^2\) and
\(\mathcal L_{\mathrm{codebook}}=\|\sg[u]-e\|_2^2\),
and \(\mathcal L_{\mathrm{usage}}\) is the normalized deficit between maximum
and empirical marginal code entropy in each product codebook, discouraging
collapse without increasing channel capacity.

\subsection{Combined objective}

After plan matching, the canonical full objective is
\begin{align}
\mathcal L
={}&1.00\mathcal L_{\mathrm{tok}}
+1.00\mathcal L_{\mathrm{plan}}
+1.00\mathcal L_{\mathrm{route}}
+0.50\mathcal L_{\mathrm{progress}}
\nonumber\\
&+1.00\mathcal L_{\mathrm{write}}
+0.25\mathcal L_{\mathrm{note}}
+0.10\mathcal L_{\mathrm{order}}
\nonumber\\
&+0.25\mathcal L_{\mathrm{commit}}
+1.00\mathcal L_{\mathrm{codebook}}
+0.10\mathcal L_{\mathrm{usage}}.
\label{eq:objective}
\end{align}
These coefficients are a registered starting configuration, not an asserted
optimum; a pre-compiled one-factor architecture screen
(Appendix~\ref{app:implementation}) varies capacity, channel, and optimizer
factors before any selected interaction is tested in a separate registered
run.

\section{Curriculum and Optimization}
\label{sec:curriculum}

Training separates capability questions that a single end-to-end run would
confound; Table~\ref{tab:curriculum} gives the four stages. Every training
example randomly maps its three semantic plans and target sections onto the
three physical decoders, with all six permutations eligible across examples:
the physical stacks may diverge numerically, but none can rely on a fixed
semantic job, and an evaluation-time plan swap must move ownership with the
plan while prompt and weights stay fixed.

\begin{table}[htbp]
\centering
\small
\begin{tabularx}{\textwidth}{
>{\raggedright\arraybackslash}p{0.08\textwidth}
>{\raggedright\arraybackslash}p{0.22\textwidth}
>{\raggedright\arraybackslash}p{0.22\textwidth}
>{\raggedright\arraybackslash}X}
\toprule
Stage & Steps and plan & Trainable components & Scientific question \\
\midrule
0 & 0--3,749; oracle outline &
Three upper decoder banks, Plan-KV, SNC, semantic heads, note writer &
Can the executor realize a correct three-way decomposition when the plan is
given? \\
1 & 3,750--9,999; learned outline &
Planner only; executor and semantic heads frozen &
Can the planner reproduce an unordered teacher plan set through a fixed
executor interface? \\
2 & 10,000--24,999; learned outline &
Planner, upper decoder banks, Plan-KV, SNC, semantic heads, note writer &
Can planning and execution adapt jointly over a frozen knowledge trunk? \\
3 & 25,000--49,999; learned outline &
Same topology as Stage 2 &
Does continued joint training improve the same mechanism without a second
path? \\
\bottomrule
\end{tabularx}
\caption{Canonical training curriculum. Stage-specific loss weights disable
objectives whose required module is frozen.}
\label{tab:curriculum}
\end{table}

The shared lower trunk remains frozen in all stages. Physical branch weights
keep FP32 optimizer masters with BF16 grouped computation; packed teacher
forcing processes one 32-token block per lane in a single physical-frontier
call, then publishes all active block notes. AdamW uses gradient clipping,
cosine decay, and stage-aware module manifests. Checkpoints bind the trunk
model and revision, fork layer, decoder count, coordination condition,
optimizer, scheduler, step, and stage, so a checkpoint cannot silently load
into a different architecture or a different bus/self-only condition.

\section{Evaluation Protocol and Current Status}
\label{sec:evaluation}

\subsection{Questions, interventions, and gates}

The evaluation is designed to prevent one success from standing in for
another. Table~\ref{tab:gates} lists the principal questions; all confidence
intervals are computed over documents rather than individual facts.

\begin{table}[htbp]
\centering
\small
\begin{tabularx}{\textwidth}{
>{\raggedright\arraybackslash}p{0.18\textwidth}
>{\raggedright\arraybackslash}p{0.25\textwidth}
>{\raggedright\arraybackslash}X}
\toprule
Question & Conditions & Preregistered evidence \\
\midrule
Executor capability &
Oracle teacher plan &
Owner-fact recall lower bound \(\geq90\%\); unauthorized-leakage upper bound
\(\leq10\%\); outline completion \(\geq90\%\); every lane long-form. \\
Planner capability &
Oracle vs.\ learned vs.\ zero plan &
Learned owner recall retains \(\geq80\%\) of oracle; zeroing all plan nodes
drops owner recall by \(\geq30\) percentage points. \\
Plan causality &
Learned plan vs.\ physical \(D_1/D_2\) plan swap &
Swapped outputs retain \(\geq80\%\) of learned-plan owner recall at moved
addresses; moved-address scoring beats deliberately unmoved scoring by
\(\geq30\) points. \\
Dynamic communication &
Bus vs.\ dynamic-note removal &
Document-bootstrap NLL change at exact dependency vs.\ nondependency tokens
and their difference; Plan-KV intact in both conditions. \\
Sibling information &
Bus vs.\ parameter-matched self-only memory &
Owner/reference realization, delayed-reference timing, duplication, and
prose quality under identical plan paths and training budgets. \\
Parallel execution &
One lane; three serial lanes; packed frontier &
Decode rounds, complete-answer latency, aggregate tokens/s, prefill and
decode time, peak memory, per-round call and cache traces. \\
\bottomrule
\end{tabularx}
\caption{Evaluation questions and preregistered evidence. Thresholds are
gates for future measurements, not current results.}
\label{tab:gates}
\end{table}

\subsection{Fact ownership and long-form quality}

Free generation stops each lane at its own EOS or at 1,000 new tokens. A
valid lane contains at least 700 tokens and four paragraphs with no more than
10\% of sentences shorter than eight words, rejecting the short-sentence
failure mode directly. Fact presence is not exact string identity: a pinned
NLI cross-encoder scores each atomic claim against sentence-level evidence in
each physical lane, with a threshold calibrated once on disjoint teacher
prose containing true and hard-negative claims, then frozen for held-out
generation. The automatic score is a screen only. Every
condition--lane--fact decision is exported to a blinded queue, judged by two
independent annotators, and resolved by a distinct third adjudicator; final
owner recall, reference recall, unauthorized leakage, and hard-negative rates
use the adjudicated judgments.

\subsection{Causal communication measurement}

Teacher-forced communication evaluation reruns the exact target with dynamic
notes removed and persistent Plan-KV unchanged. Retokenization has already
marked every token overlapping a required receiver quote and verified that
its source block is visible under the one-block delay. For token \(q\),
define
\begin{equation}
\Delta_q
=\operatorname{NLL}_{\mathrm{no\text{-}dynamic}}(q)
-\operatorname{NLL}_{\mathrm{bus}}(q).
\end{equation}
The evaluator reports the document-level mean \(\Delta\) on dependency
tokens, on all other active tokens, and their selectivity difference: the
mechanism predicts a larger benefit at annotated delayed dependencies, so a
global likelihood change is insufficient. Destructive controls---mutating one
producer's code, exchanging note sources, scrambling note directions at
matched norm, closing the note gate---diagnose address and payload
sensitivity but do not replace the primary bus-versus-self-only comparison.

\subsection{Scale control and status}

The first model rung is a dense Qwen3-4B checkpoint \citep{yang2025qwen3}. If
its oracle-plan executor fails, data alignment, ownership labels, routing
heads, optimization, cache behavior, and plan delivery are audited first;
only if those pass is the same architecture repeated with a dense 14B trunk,
asking whether the negative result is a trunk-capacity result rather than
changing the scientific condition.

At the time of writing, the three physical decoder stacks, private caches,
planner, Plan-KV, product-VQ notes bus, packed training rollout,
counterfactual runtime, historical source filter, strict teacher schemas, and
evaluators are implemented, and four manually audited long-form records
exercise the data and trainer contracts, including sparse and
zero-dependency cases. These establish executable plumbing only; no quality,
coordination, or speedup result is reported in this manuscript.

\section{Related Work}
\label{sec:related}

\paragraph{Prompted and learned semantic branching.}
Skeleton-of-Thought expands an elicited textual skeleton through parallel
requests or batched decoding \citep{ning2024sot}; APAR post-trains fork
control tokens and restricts attention over the paragraph tree
\citep{liu2024apar}; PASTA learns a promise-and-sync annotation language so
independent chunks decode asynchronously \citep{jin2025promise}. \pdt\ shares
the premise that reading order need not equal generation order, but makes
three persistent upper decoder stacks part of the model, each conditioned on
a continuous plan and a finite delayed latent channel.

\paragraph{Concurrent workers with high-bandwidth visibility.}
Hogwild! Inference lets concurrent model instances attend through a
concurrently updated shared cache \citep{rodionov2025hogwild}; Group Think
gives concurrent reasoning trajectories token-level mutual visibility
\citep{hsu2025groupthink}. \pdt\ studies the opposite communication regime:
private causal caches, an explicit one-block barrier, and 32 learned payload
bits per lane block, which makes causal attribution and channel capacity
directly measurable.

\paragraph{Non-autoregressive and multi-token generation.}
Non-autoregressive translation predicts many target positions in parallel,
trading away the left-to-right factorization and often relying on
distillation \citep{gu2018nat,kim2016seqkd}. Speculative decoding accelerates
sampling while preserving the target distribution \citep{leviathan2023spec},
Medusa adds multiple decoding heads \citep{cai2024medusa}, EAGLE speculates
in feature space \citep{li2024eagle}, and multi-token prediction trains
independent future-token heads above a shared trunk
\citep{gloeckle2024multitoken}. \pdt\ instead gives each head a complete,
persistent autoregressive section that intentionally differs from the
sequence a single left-to-right decoder would have sampled.

\paragraph{Set prediction, discrete latents, and external memory.}
Attention over learned seed queries yields set-valued outputs
\citep{lee2019settransformer}; product vector quantization supplies a
differentiable finite alphabet \citep{vandenOord2017vq};
retrieval-augmented generation couples a parametric generator to external
non-parametric memory \citep{lewis2020rag}. \pdt's plan and notes are
internal, document-specific state derived from the current source and
generated blocks.

\section{Discussion and Limitations}
\label{sec:discussion}

\paragraph{Physical decoders without constant communication.}
The notes bus need not fire on every example for the physical split to
matter. Three independent upper stacks can adapt to concurrent long-form
trajectories, maintain private causal summaries, and follow different
persistent plans without forcing all three roles through one upper
parameterization; the model still emits three tokens per round when the
optimal message is uninformative. Random semantic-to-physical assignment
keeps this from collapsing into fixed experts: each stack may learn
\emph{trajectory capacity}, while its subject matter comes from the assigned
plan. The self-only and no-dynamic controls price the physical
factorization; the bus ablation prices sibling communication alone.

\paragraph{Persistent attention rather than a latent seed.}
A single additive vector can be ignored, overwritten, or entangled with lane
state. Plan-KV instead exposes each outline node as addressable memory at
every upper layer and token, asking the model to learn queries into a
persistent structured memory rather than assuming instructions correspond to
a linear direction in the residual stream. Complete-plan swaps then test
causality: if assignment is real, ownership must move while physical weights
stay fixed.

\paragraph{Offline and fixed-corpus deployment.}
Because the planner and notes bus operate entirely on local hidden states,
the architecture is compatible with disconnected or fixed-corpus systems
(onboard manuals, operating procedures, emergency protocols) where an online
retrieval service \citep{lewis2020rag} is unavailable. This is a deployment
hypothesis, not evidence of transfer: safety-critical use would require
domain data, expert validation, failure-mode testing, and a separate safety
case.

\paragraph{Falsifiability.}
The design is weakened by any of the following outcomes:
\begin{itemize}[leftmargin=*,itemsep=0.2em]
    \item the oracle-plan executor cannot achieve reliable lane ownership:
    three physical stacks do not by themselves solve assignment;
    \item the learned planner stays far below a sufficient oracle executor:
    decomposition is the bottleneck;
    \item plan swaps leave facts at their original physical addresses:
    branches learned fixed roles or ignore Plan-KV;
    \item dynamic-note removal has no selective effect on annotated
    dependency tokens: the bus is unused or unnecessary;
    \item the self-only control matches the bus on dependency-sensitive
    held-out data: sibling communication adds no measured value;
    \item grouped execution fails to reduce practical complete-answer
    latency: algorithmic span is not realized on the tested hardware.
\end{itemize}
These outcomes are separable: a failed notes bus does not erase the physical
parallel frontiers, but it falsifies the stronger claim that a finite latent
channel supports useful cross-lane coordination.

\paragraph{Limitations.}
Three independent upper stacks increase parameters, optimizer state, and
upper-layer memory traffic; the canonical 24/12 split is a test point, not
proof that half-depth branching is optimal; and Equation~\ref{eq:span} must
not be quoted as measured latency. The initial evidence domain is English
historical exposition, chosen because facts, chronology, citations, and
multi-section structure can be audited---code, mathematics, fiction, and
tightly sequential reasoning may admit different decompositions or none at
all. Wikipedia assessments and citations are filters, not guarantees;
teacher-generated facts and plans can introduce errors under exact
provenance schemas; and BGE targets, NLI screens, and product-VQ codes are
model-dependent representations, which is why blinded human adjudication
remains in the protocol. Finally, the one-block barrier excludes useful
same-round interaction by design, and a 32-bit message may be too narrow,
while a wider or more frequent channel may erase the constraint under study;
the architecture screen measures this trade-off but cannot remove it.

\section{Conclusion}
\label{sec:conclusion}

The Parallel Decoder Transformer turns long-form parallel generation into a
model architecture rather than an external scheduling trick. A frozen shared
lower trunk supplies common language and world knowledge; a continuous
planner produces an unordered set of three outlines; three independently
parameterized upper stacks own private causal frontiers; persistent Plan-KV
supplies structural assignment; and a finite delayed notes bus provides the
only lawful path for cross-lane influence---three tokens per synchronization
round without violating autoregression inside any section. Whether the
architecture learns the intended behavior is the empirical question the
registered protocol separates into executor capability, planner
decomposition, plan causality, and dynamic communication; those experiments
are in progress. The contribution of this paper is the precise,
implementable, falsifiable design against which the eventual result can be
judged.

\appendix

\section{Canonical Instantiation and Registered Training Configuration}
\label{app:implementation}

The first implementation uses \texttt{Qwen/Qwen3-4B-Instruct-2507} at a
pinned revision \citep{yang2025qwen3}: 36 decoder layers, hidden width
2,560, layers 0--23 frozen shared, layers 24--35 initializing each of the
three physical upper stacks. Table~\ref{tab:implementation} records the
canonical dimensions. The physical stack keeps FP32 master parameters with
BF16 grouped compute views; frozen lower-layer execution is outside
autograd, and exact causal grouped-query attention is evaluated in bounded
query tiles with activation recomputation.

\begin{table}[htbp]
\centering
\small
\begin{tabular}{lr}
\toprule
Component & Canonical value \\
\midrule
Physical decoder lanes & 3 \\
Shared / physical upper layers & 24 / \(3\times12\) \\
Planner nodes per lane & 8 \\
Planner width / layers / heads & 512 / 2 / 8 \\
Plan and notes memory width & 256 \\
Plan/SNC attention width / heads & 512 / 8 \\
Target block size & 32 tokens \\
Dynamic codebooks / entries & 4 / 256 \\
Logical dynamic capacity & 32 bits per lane block \\
Dynamic history & 16 blocks \\
Plan and notes outer gate initialization & \(-4.0\) \\
Shared frozen parameters & 2,811,298,304 \\
Physical upper-stack parameters & 3,633,509,376 \\
Planner and coordination parameters & 84,132,905 \\
Total materialized parameters & 6,528,940,585 \\
\bottomrule
\end{tabular}
\caption{Canonical architecture configuration. Parameter counts are an
implementation census, not a performance result.}
\label{tab:implementation}
\end{table}

The registered optimizer is AdamW with learning rate \(2\times10^{-4}\),
weight decay 0.01, 1,250 warmup steps, cosine decay, gradient-norm clipping
at 1.0, batch size one document, and gradient accumulation over 16
documents. The 16,384-token planner-prompt maximum is a fail-fast ceiling;
source documents are never truncated to meet it. The architecture screen
contains 25 one-factor variants at three independent seeds, varying model
scale, upper-stack depth, planner depth and feed-forward width, notes and
attention width, product-code shape, gate initialization, learning rate,
weight decay, and bus versus parameter-matched self-only memory; it is
compiled before training, and any interaction study is registered after the
screen rather than selected post hoc.

\bibliographystyle{plainnat}
\bibliography{refs}

@inproceedings{ning2024sot,
  title     = {Skeleton-of-Thought: Prompting {LLM}s for Efficient Parallel Generation},
  author    = {Ning, Xuefei and Lin, Zinan and Zhou, Zixuan and Wang, Zifu and Yang, Huazhong and Wang, Yu},
  booktitle = {International Conference on Learning Representations},
  year      = {2024},
  url       = {https://arxiv.org/abs/2307.15337}
}

@misc{liu2024apar,
  title         = {{APAR}: {LLM}s Can Do Auto-Parallel Auto-Regressive Decoding},
  author        = {Liu, Mingdao and Zeng, Aohan and Wang, Bowen and Zhang, Peng and Tang, Jie and Dong, Yuxiao},
  year          = {2024},
  eprint        = {2401.06761},
  archivePrefix = {arXiv},
  primaryClass  = {cs.CL},
  url           = {https://arxiv.org/abs/2401.06761}
}

@inproceedings{jin2025promise,
  title     = {Learning to Keep a Promise: Scaling Language Model Decoding Parallelism with Learned Asynchronous Decoding},
  author    = {Jin, Tian and Cheng, Ellie Y. and Ankner, Zack and Saunshi, Nikunj and Elias, Blake M. and Yazdanbakhsh, Amir and Ragan-Kelley, Jonathan and Subramanian, Suvinay and Carbin, Michael},
  booktitle = {International Conference on Machine Learning},
  year      = {2025},
  url       = {https://arxiv.org/abs/2502.11517}
}

@misc{rodionov2025hogwild,
  title         = {Hogwild! Inference: Parallel {LLM} Generation via Concurrent Attention},
  author        = {Rodionov, Gleb and Garipov, Roman and Shutova, Alina and Yakushev, George and Egiazarian, Vage and Sinitsin, Anton and Kuznedelev, Denis and Alistarh, Dan},
  year          = {2025},
  eprint        = {2504.06261},
  archivePrefix = {arXiv},
  primaryClass  = {cs.LG},
  url           = {https://arxiv.org/abs/2504.06261}
}

@misc{hsu2025groupthink,
  title         = {Group Think: Multiple Concurrent Reasoning Agents Collaborating at Token Level Granularity},
  author        = {Hsu, Chan-Jan and Buffelli, Davide and McGowan, Jamie and Liao, Feng-Ting and Chen, Yi-Chang and Vakili, Sattar and Shiu, Da-shan},
  year          = {2025},
  eprint        = {2505.11107},
  archivePrefix = {arXiv},
  primaryClass  = {cs.CL},
  url           = {https://arxiv.org/abs/2505.11107}
}

@inproceedings{gu2018nat,
  title     = {Non-Autoregressive Neural Machine Translation},
  author    = {Gu, Jiatao and Bradbury, James and Xiong, Caiming and Li, Victor O. K. and Socher, Richard},
  booktitle = {International Conference on Learning Representations},
  year      = {2018},
  url       = {https://arxiv.org/abs/1711.02281}
}

@inproceedings{kim2016seqkd,
  title     = {Sequence-Level Knowledge Distillation},
  author    = {Kim, Yoon and Rush, Alexander M.},
  booktitle = {Conference on Empirical Methods in Natural Language Processing},
  year      = {2016},
  url       = {https://aclanthology.org/D16-1139/}
}

@inproceedings{leviathan2023spec,
  title     = {Fast Inference from Transformers via Speculative Decoding},
  author    = {Leviathan, Yaniv and Kalman, Matan and Matias, Yossi},
  booktitle = {International Conference on Machine Learning},
  year      = {2023},
  url       = {https://arxiv.org/abs/2211.17192}
}

@inproceedings{cai2024medusa,
  title     = {Medusa: Simple {LLM} Inference Acceleration Framework with Multiple Decoding Heads},
  author    = {Cai, Tianle and Li, Yuhong and Geng, Zhengyang and Peng, Hongwu and Lee, Jason D. and Chen, Deming and Dao, Tri},
  booktitle = {International Conference on Machine Learning},
  year      = {2024},
  url       = {https://arxiv.org/abs/2401.10774}
}

@inproceedings{li2024eagle,
  title     = {{EAGLE}: Speculative Sampling Requires Rethinking Feature Uncertainty},
  author    = {Li, Yuhui and Wei, Fangyun and Zhang, Chao and Zhang, Hongyang},
  booktitle = {International Conference on Machine Learning},
  year      = {2024},
  url       = {https://arxiv.org/abs/2401.15077}
}

@inproceedings{gloeckle2024multitoken,
  title     = {Better \& Faster Large Language Models via Multi-token Prediction},
  author    = {Gloeckle, Fabian and Idrissi, Badr Youbi and Rozi{\`e}re, Baptiste and Lopez-Paz, David and Synnaeve, Gabriel},
  booktitle = {International Conference on Machine Learning},
  year      = {2024},
  url       = {https://arxiv.org/abs/2404.19737}
}

@inproceedings{lee2019settransformer,
  title     = {Set Transformer: A Framework for Attention-based Permutation-Invariant Neural Networks},
  author    = {Lee, Juho and Lee, Yoonho and Kim, Jungtaek and Kosiorek, Adam R. and Choi, Seungjin and Teh, Yee Whye},
  booktitle = {International Conference on Machine Learning},
  year      = {2019},
  url       = {https://arxiv.org/abs/1810.00825}
}

@inproceedings{vandenOord2017vq,
  title     = {Neural Discrete Representation Learning},
  author    = {van den Oord, Aaron and Vinyals, Oriol and Kavukcuoglu, Koray},
  booktitle = {Advances in Neural Information Processing Systems},
  year      = {2017},
  url       = {https://arxiv.org/abs/1711.00937}
}

@inproceedings{lewis2020rag,
  title     = {Retrieval-Augmented Generation for Knowledge-Intensive {NLP} Tasks},
  author    = {Lewis, Patrick and Perez, Ethan and Piktus, Aleksandra and Petroni, Fabio and Karpukhin, Vladimir and Goyal, Naman and K{\"u}ttler, Heinrich and Lewis, Mike and Yih, Wen-tau and Rockt{\"a}schel, Tim and Riedel, Sebastian and Kiela, Douwe},
  booktitle = {Advances in Neural Information Processing Systems},
  year      = {2020},
  url       = {https://arxiv.org/abs/2005.11401}
}

@misc{xiao2023cpack,
  title         = {{C-Pack}: Packed Resources for General Chinese Embeddings},
  author        = {Xiao, Shitao and Liu, Zheng and Zhang, Peitian and Muennighoff, Niklas and Lian, Defu and Nie, Jian-Yun},
  year          = {2023},
  eprint        = {2309.07597},
  archivePrefix = {arXiv},
  primaryClass  = {cs.CL},
  url           = {https://arxiv.org/abs/2309.07597}
}

@misc{yang2025qwen3,
  title         = {{Qwen3} Technical Report},
  author        = {{Qwen Team}},
  year          = {2025},
  eprint        = {2505.09388},
  archivePrefix = {arXiv},
  primaryClass  = {cs.CL},
  url           = {https://arxiv.org/abs/2505.09388}
}

\end{document}